\documentclass[11pt]{article}
\usepackage[utf8]{inputenc}
\usepackage{amsmath, amssymb}
\usepackage[a4paper, margin=1in]{geometry}
\usepackage{setspace}
\usepackage{newpxtext}
\usepackage{authblk}
\usepackage{graphicx}

\title{The Representational Status of Deep Learning Models}

\author[1,2]{Eamon Duede\thanks{Email: eduede@purdue.edu}}
\affil[1]{Purdue University}
\affil[2]{Argonne National Laboratory}

\date{} 

\begin{document}

\maketitle

\section{Introductory}

It is commonplace for scientists and philosophers alike to refer to fully trained deep learning architectures as `representations' or `models'. However, this interchangeable usage can lead to confusion, as the specific meanings behind these terms can vary depending on the field and context of use. What is meant by `representation' or `model' in one context might be rather incoherent in another. For example, when AI researchers examine how `representations' in a deep network are affected by elements of that network's architecture, what counts as a representation here is the observed `pattern of neural activation' attributable to that architectural element in concert with neural inputs \cite{olsson2022context}. In this context, `representation' has much in common with its use in neuroscience, where neural activation patterns are viewed as \textit{functional} \cite{cao2022putting}. Understood functionally, representations are mechanisms that have identifiable effects on behavior or serve to facilitate the accomplishment of some goal \cite{dretske1981knowledge,dretske1991explaining,millikan1989biosemantics,neander1995misrepresenting}. Of course, considerable ambiguity remains within the AI and neuroscience communities on how best to spell out that functional conception \cite{baker2022three}.

While the AI and neuroscience literatures have a tendency to focus on a functional conception of representation, the philosophical literature on scientific models focuses on a \textit{relational} conception. Here, models stand in a representational relationship to their targets. In general, this has been understood to mean that features of the model are either similar to, correspond with, denote, or otherwise exemplify relevant aspects of the model's target system. If some such relation or set of relations holds, the model is said to represent its target. However, as with the functional conception, philosophers remain divided on how best to cash out the fine details of the relational conception. Nevertheless, while there is undoubtedly scope for some overlap,\footnote{For instance, the brain may represent targets both functionally and relationally \cite{shagrir2018brain}, and, moreover, some plausible readings of Dretske would have him attempting to naturalize relational representation by way of functional representation.} in general, the functional and relational conceptions of representation appear meaningfully distinct.

The aim of this chapter is, then, to gain clarity on the otherwise overlooked question of the representational status of deep learning models (DLMs) understood in the way implied by the relational conception. Viewed from the functional perspective of systems neuroscience or fundamental AI research, the question of whether deep learning models are representational is a trivial one ---of course they are.\footnote{It would be a miracle if not \cite{rowbottom2024does}.} But, this is to misunderstand the question posed by the present study. Considered from the relational perspective, the representational status of deep learning models remains unresolved. 

The position I will argue for and defend throughout the chapter is that deep learning models do, indeed, represent their targets in the relational sense. However, I show that, in general, we have no good reason to believe that DLMs encode what I will call locally semantically decomposable representations of their targets. That is, the available evidence does not warrant a general belief that a deep learning model's units, sets or layers of units, and associated weights represent individual aspects of the target system \textit{relationally}. Rather, scientists are, at most, justified in ascribing to DLMs a representational status akin to that of scientific models whose representational power hinges on their holistic properties (for instance, in a way similar to that of, say, the Ising model). That is, I will argue that the representational capacity these models have is largely global, rather than decomposable into stable, local sub‐representations. This result has immediate epistemic consequences for interpretability approaches in the budding field of explainable AI (XAI). Additionally, this analysis helps to redirect philosophical attention towards exploring the global relational nature of deep learning representations and their relationship both to models more generally and their potential role in future scientific inquiry.

There are at least two main reasons why the representational status of deep learning models has been neglected. The first is an effect due to ambiguity in the concept of representation in philosophy \cite{batterman2014minimal,de2021representationalism} and in neuroscience \cite{baker2021philosophical,cao2018computational}. Historically, these fields have had a tendency to carelessly conflate functional and relational notions of representation. The broad, conceptual and historical influence these literatures have had on contemporary research in AI \cite{harvey2008misrepresentations,cao2022putting} has, in turn, led to confusion in the AI literature \cite{yarkoni2019generalizability,shanahan2022talking,mitchell2021ai,shevlin2019apply}. Indeed, as I will show throughout, not only are deep learning models routinely referred to as `representations' in the AI literature, but their architectural components are also described using language that has the effect of schooling the eye to mistakenly see DLMs as more richly and relationally representational than is currently justified. Second, a general lack of model transparency \cite{creel2020transparency} makes it difficult to directly determine the representational status of DLMs, allowing certain conceptions and misconceptions to become entrenched. 

Resolving the representational status of deep learning is not merely an exercise in the metaphysics of AI. Nor is it philosophical nitpicking of careless AI researchers. The issue has immediate epistemological and ethical consequences. For instance, many (though not all) current model interpretability efforts appear to \textit{assume} a richer representational status than has yet been established, rendering such approaches highly under-determined, unfalsifiable, and likely spurious \cite{leavitt2020towards}. This can lead, in turn, to unjustified beliefs about model reliability \cite{duede2022instruments}, targets \cite{rudin2019stop}, or unethical decision-making practices where accountability and value-alignment are salient \cite{birch2022clinical, falco2021governing, hoffman2017taxonomy}.

Going forward, I will use the term `representation$_\text{\textbf{rel}}$ to refer to that which stands in a representational relation, as understood on a \textit{relational} conception. I will, of course, have occasion to discuss representations from the functional perspective when, for instance, I discuss neural activation patterns in deep neural networks and human brains in \textit{Section 4}. In such cases, I will clarify that I mean by `representation' the sense denoted by the functional perspective by using `representation$_\text{func}$'. Nevertheless, I will, for the most part, instead adopt the term `learned transformation' to avoid confusion.

Here, then, is the plan: In \textit{Section 2}, focusing on the role of representation$_\text{rel}$, I briefly review the recent literature on scientific models to bring out a novel graded dimension of semantic decomposability. On one end are models that are `locally semantically decomposable', and on the other, those that are globally or `holistically decomposable'. I demonstrate that this dimension is orthogonal to the degree to which any given model is idealized. In \textit{Section 3}, I turn to a discussion of deep learning models, including their structure and training process, as well as the challenges of resolving their representational$_\text{rel}$ status presented by opacity. In doing so, I clarify relevant terminology and adopt a stable vocabulary to avoid ambiguity. In \textit{Section 4}, I examine whether more technical, conceptual, and philosophical considerations support the idea that DLMs encode locally decomposable representations$_\text{rel}$. I conclude in \textit{Section 5} that DLMs are representational$_\text{rel}$ in a global or holistic sense and that there is no evidence that they are more locally decomposable models of their targets, thereby highlighting potential epistemic and practical risks associated with interpreting them as such. At the same time, this analysis reveals that, due to opacity, it is not possible to assess the degree of idealization for any particular model.\footnote{Though there may be some reason to think that some deep learning models are more idealized than others as a function of the dimensionality of their inputs (for instance, representing global weather patterns as a function of two input parameters implies a high degree of idealization while doing so with a million suggests less idealization).}

\section{Representational Scientific Models}

The question that concerns us in this chapter is not whether DLMs are \textit{models}. While there is certainly debate concerned with which things count as models, for the purposes of the arguments to follow, I simply assert that scientific models are whichever things scientists take to be or use as models.\footnote{This view is not without substantial motivation and support from empirical, conceptual, and historical evaluations of scientific practice \cite{plutynski2001modeling,parker2020model,giere2010explaining}. It should also be noted that if you are inclined to argue that DLMs are not models, then you are likely already disposed to agree that they are not representational$_\text{rel}$, in which case your position is, in fact, stronger than mine.} Instead, this chapter seeks to understand whether and to what extent deep learning models are \textit{representational} models understood from the \textit{relational perspective}.

There is a large and diverse literature that argues that not all models are faithfully representational \cite{wimsatt1987false,van2010scientific}; that models merely denote targets or exemplify relevant properties \cite{frigg2010fiction,frigg2016fiction}; that models are essentially tools and not necessarily representations \cite{giere2004models,parker2020model,boon2009models}; that what makes something a model is the functions it serves regardless of its representational status \cite{suarez2004inferential,contessa2007scientific}; that the appropriate epistemic evaluation of a model is a function of adequacy \cite{parker2020model} and not its similarity to targets --on and on. What gives rise to and motivates much of the diversity in this literature is a concern to make sense of how models can give us knowledge about the world in cases where they explicitly \textit{misrepresent} the world. Of course, by necessity, all instances of \textit{misrepresntation} are cases of representation$_\text{rel}$, since there is \textit{something} that is being represented, just badly. After all, it is well established that most models \textit{misrepresent} their targets \cite{cartwright1984laws,batterman2010explanatory,rice2012optimality} and, for example, the more highly idealized the model, the less representationally$_\text{rel}$ faithful it is. Nevertheless, it has also been argued that, in the extreme case, so-called `targetless' models do not represent$_\text{rel}$ any real, nomologically possible, targets at all \cite{frigg2020modelling,weisberg2012simulation,levy2015modeling,massimi2019two}.\footnote{But, see \cite{verreaultrepresenting} for a recent argument that there are no `targetless' models.} Regardless, it is fairly straightforward to show that deep learning models have targets (see \textit{Section 3}).

How, exactly, to cash out the nature of the representational relation that is thought to obtain in cases of representational$_\text{rel}$ modeling is dicey. While some scholars have advocated for technical specifications such as relations of isomorphism \cite{french2003model,van1980scientific,da2003science}, what most every representational$_\text{rel}$ account of scientific models turns on are notions or requirements of `similarity' and `correspondence' between the model and the target \cite{giere2010explaining,weisberg2012simulation,baker2021philosophical}.\footnote{Though many philosophers have pointed out that similarity-based views fail in corner cases and give rise to unresolvable logical, ontological, and epistemic problems \cite{wimsatt1987false,parker2020model,sua2003scientific}.} Of course, similarity admits of degrees and, moreover, there are a variety of ways in which model and target can be similar. After all, a physical scale model of a bridge can be more or less similar to its target. For instance, it might have the same number of components. But, its similarity would increase by having the same relative positions and size ratios of those components. And, more similar still by using the same materials --and so on, until it is an exact Borges replica.

When it comes to representation$_\text{rel}$, an essential aspect of similarity is the degree of model idealization. A model's degree of idealization picks out how much it simplifies, omits, or distorts aspects of the real system. But, consideration of highly idealized models reveals that meaningful distinctions concerning target similarity exist between them that are not picked out by degree of idealization.

To see this, first consider the Lotka-Volterra model used in population biology: 

\begin{equation}
\begin{aligned}
& \frac{d x}{d t}=\alpha x-\beta x y \\
& \frac{d y}{d t}=-\gamma y+\delta x y
\end{aligned}
\end{equation}

The Lotka-Volterra model is an example of a highly idealized model. The model's degree of idealization results from a number of key assumptions that are built into its mathematical structure. For one thing, the model is set on the reals, resulting in continuous, non-integer values for quantities that are not measurable in anything but integers (e.g., population size or number of interactions). Additionally, the model assumes a fixed interaction rate between predators and prey, with no alternative food sources or adaptive behaviors. It neglects intra-species competition, treating populations as homogeneous with no age structure or genetic variability. Birth and death rates are fixed, ignoring seasonal fluctuations or environmental changes. Furthermore, the model is entirely deterministic, assuming smooth population changes without stochastic effects. Spatial structure is also absent, with populations assumed to be well-mixed rather than distributed across heterogeneous environments.

Surprisingly, however, the Lotka-Volterra model is relevantly similar in representational character to that of the scale model of a bridge. Like in the case of the bridge, one can specify in a regular and systematic manner what all of the constituents of the Lotka-Volterra model \textit{mean} in terms of the target system, as well as how the structure or composition of those constituents contributes to that meaning. For instance, in the Lotka-Volterra model, $x(t)$ and $y(t)$ correspond to the sizes of the prey and predator populations, respectively, at some time. $\alpha$ corresponds to the prey birth rate, and multiplying $x$ with this rate at each time-step causes prey to reproduce exponentially in the absence of predators. $\beta$ is the predation rate (a probability that a predator-prey encounter leads to prey being consumed). At the same time, the second equation governs interaction and predator population dynamics. $\gamma$ is the predator death rate (predators die at a constant rate in the absence of prey), and $\delta$ is the predator reproduction efficiency (how well consumed prey are converted into new predators). The model’s structure and composition (the syntax of the model) capture relations of dependence or correlation between elements or parameters (its semantics) in the model in a way that also maps to the world (e.g., predators and prey interact $\beta x y$). 

In this way, we can say of the Lotka-Volterra model that it is \textit{both} highly idealized \textit{and} highly (locally) semantically decomposable (though from here on I will simply say `highly decomposable'). Consideration of the Lotka-Volterra model reveals that when it comes to representation$_\text{rel}$, in addition to degree of idealization, another essential aspect of similarity is the degree to which salient interpretations in terms of the target can be assigned to constituent elements of the model. In some models (like the bridge or the Lotka-Volterra), the majority of parameters or structural components can each be given a stable semantic meaning that directly corresponds to an aspect of the target system. Yet, not all models are like this. In what I refer to as holistic or globally semantically decomposable models (or simply `minimally decomposable'), the representational capacity resides in the aggregate structure of the model rather than in individually interpretable parts. In that case, one \textit{cannot} systematically say of this or that subcomponent of the model that it stands for this or that aspect of the target, even if the model as a whole still stands in a meaningful representational$_\text{rel}$ relationship to that target. For instance, a model’s structure and composition (its syntax) might encode relations of dependence or correlation between elements (its semantics) without any straightforward, one-to-one mapping between single parameters and individual real-world factors.

Moreover, the degree to which a model is semantically decomposable is unrelated to its degree of idealization. This is readily observable in the comparison of the scale model and the Lotka-Volterra model, where the former is minimally idealized and highly decomposable while the latter is highly idealized though still highly decomposable. To see that a model can also be highly idealized and minimally decomposable, consider, for instance, the Ising Model used to study phase transitions, magnetism, and statistical mechanics:

\begin{equation}
H=-J \sum_{\langle i, j\rangle} S_i S_j-h \sum_i S_i
\end{equation}

Like the Lotka-Volterra model, the Ising model is also highly idealized in several key ways. It represents spins as having only two discrete states, $\pm1$, ignoring the continuous nature of real magnetic moments. Interactions are restricted to nearest neighbors, excluding the long-range effects present in real materials. Furthermore, the model treats the system classically, disregarding quantum mechanical effects that influence real spin systems. It assumes a perfect, homogeneous, and periodic lattice, omitting structural irregularities or defects commonly found in actual materials. Thermal effects are simplified using classical Boltzmann weighting without accounting for quantum fluctuations. Additionally, external perturbations are limited to a simple uniform field $h$, neglecting more complex environmental influences that can shape magnetic behavior.

However, unlike the Lotka-Volterra model, the Ising model is not locally semantically decomposable precisely because the meaning of its constituent elements is not independently interpretable in isolation but instead emerges from the statistical behavior of the system as a whole. Each spin variable, $\sigma_i \in \{\pm1\}$, lacks an intrinsic, self-contained meaning outside of its role within the collective state of the system. The Hamiltonian $H$ does not specify distinct causal contributions of individual terms but rather encodes an energy function that governs system behavior in an aggregate manner. The interaction term, $-J \sum_{\langle i,j \rangle} \sigma_i \sigma_j$, enforces a coupling between neighboring spins, yet the effect of any particular $\sigma_i$ cannot be meaningfully understood in isolation because its contribution to the total energy depends entirely on the configuration of its neighbors. Likewise, the external field term, $-h \sum_i \sigma_i$, applies a uniform influence across all spins, but its impact is a function of the entire spin distribution rather than any single spin’s behavior. Moreover, thermodynamic properties of the model only acquire meaning through statistical averaging over all sites in the system and cannot be traced back to a single term or spin in a straightforward, interpretable way.\footnote{See \cite{rice2019models} for an extended and quite thorough treatment of this issue.}

For some highly idealized models, therefore, it is not possible to assign a semantic interpretation in terms of specific aspects of the target to either a large number of its constituents, its structure and composition, or both. In these cases, the model can still be used to explain the behavior of some broad class of phenomena, but not because it accurately represents$_\text{rel}$ its target \cite{batterman2014minimal,lange2015minimal}.\footnote{For \cite{batterman2014minimal}, similarity relations do not play a role at all for \textit{explanation}. Rather, the model's membership in a shared universality class with the target does the explaining. Yet, see \cite{lange2015minimal} for an argument that this amounts to a similarity relation after all. Regardless of the fine details, there is no debate as to whether minimal models represent targets, but rather, \textit{how they explain}. In this paper, I am concerned simply with degree of representational similarity.}

Nevertheless, highly idealized models \textit{are} representational in so far as they have real targets. In the case of the Ising model, the model is used to represent aggregate patterns in spin interactions and phase transitions and is often deployed as a computationally tractable surrogate for otherwise intractable simulations of many-body statistical systems, such as those governed by quantum or mean-field theories of magnetism. However, if one were to examine the Ising as if it were highly decomposable, then one would be liable to draw inferences concerning the physical dependencies, constitution, and so on, of actual systems that are not, in fact, represented in the model at all.

\begin{figure}[htbp]
    \medskip
    \centering
    \includegraphics[width=0.8\linewidth]{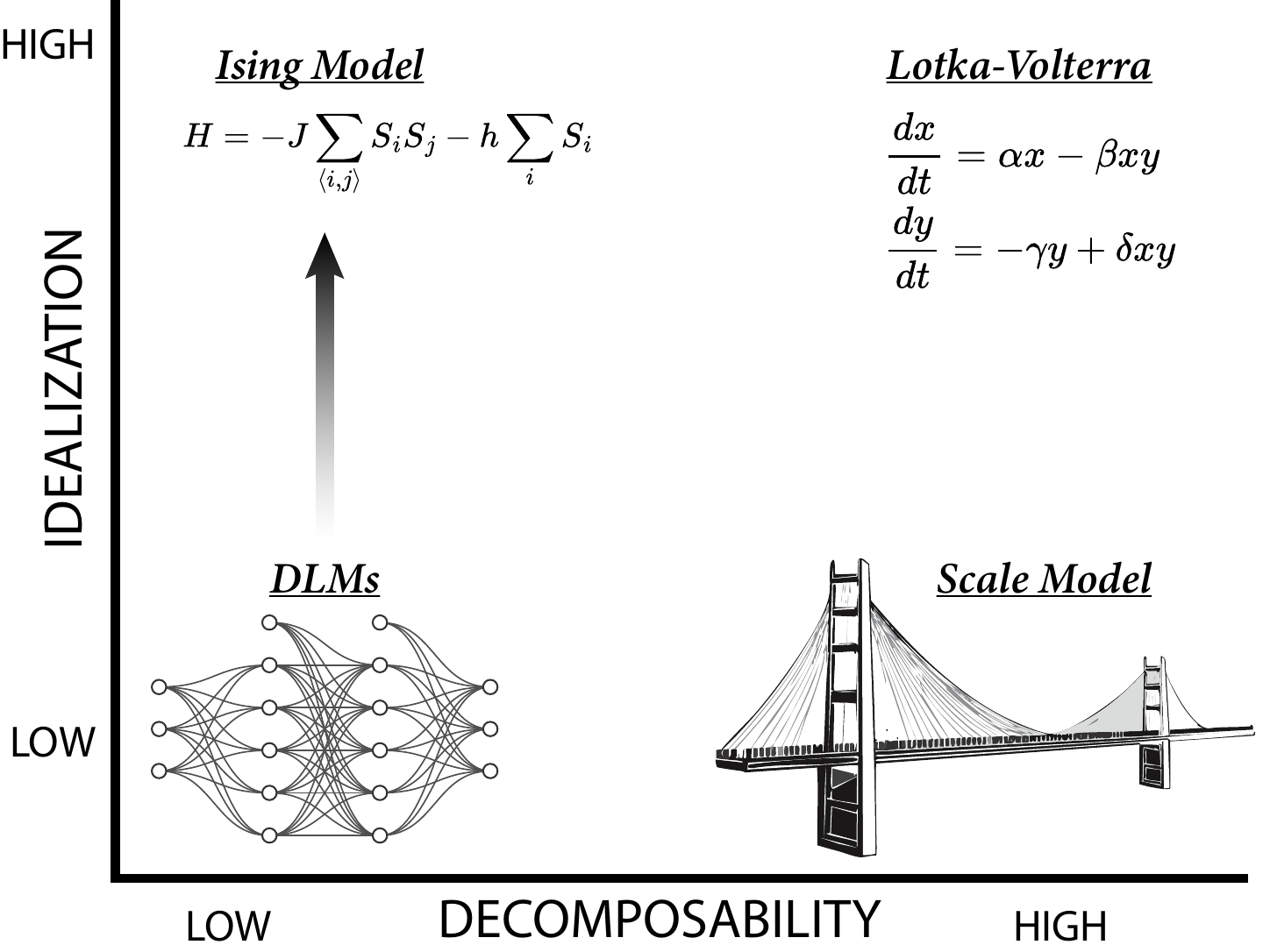}
    \caption{Degree of idealization and degree of semantic decomposability each form a separate dimension of model-building. Some models, like a scale model, are highly decomposable and minimally idealized. Others, like the Ising model, are minimally decomposable and highly idealized. Deep learning models are minimally decomposable, but, for any given model, it is not possible to determine its degree of idealization due to opacity.}
    \label{fig:case2_benfits}
    \medskip
\end{figure}

In attempting to determine the representational$_\text{rel}$ status of deep learning models, we must first ask whether they are representational$_\text{rel}$ at all. That is, do DLMs have real targets? As I will show in \textit{Section 3}, the answer to this question is a straightforward `yes'. To say that a DLM is representational$_\text{rel}$ is to place it somewhere on a graded plane, from minimally idealized to highly idealized and from minimally decomposable to highly decomposable. In \textit{Section 4}, I will show that DLMs are best characterized as minimally semantically decomposable models whose representational$_\text{rel}$ capacities are global or holistic. However, due to opacity, it is not possible to evaluate the degree of idealization for any given deep learning model. However, given the wide variance in DLM dimensionality, it is reasonable to suspect that some models are more idealized than others (see \ref{fig:case2_benfits}). 

\section{Deep Learning and Representational Ambiguity}
Deep learning is a machine learning approach that utilizes artificial neural networks \cite{lecun2015deep,goodfellow2016deep}. In machine learning, the output of running a `learning algorithm' on a training dataset $\mathcal{D}$ is often referred to as a `model'. The `model' is an estimated function $\hat{f}$ that takes inputs $X$ and maps them to outputs $Y$. Inputs and outputs are known as `features' or `attributes' and are vectors of values in some domain such as real numbers, integers, ordinals, Booleans --on and on. For now, in order to avoid confusion, I will use the term `estimator' to refer to $\hat{f}$. For most standard deep learning applications, the aim is to find an estimator that most accurately approximates the true function $f^*$ that is responsible for generating $\mathcal{D}$.

For simplicity, I will consider traditional, feed-forward deep neural networks trained in a supervised manner \cite{flach2012machine}.\footnote{What follows generalizes with some modification to almost all of the common deep learning approaches in use today. The main claims of this paper do not depend on DL approaches being fully feed-forward, and are unchanged when considering transformer architectures.} In a fully connected deep neural network architecture, each unit in a given layer is connected to every unit in the subsequent layer. The value of the output of a given unit is a function of the weighted sum of all outputs of all units from the previous layer (which are inputs to the unit).\footnote{Let $a_n^l$ denote the output of the $n^{th}$ unit in the $l_{th}$ layer such that $a_n^l = g(\sum_m w_{nm}^l a_m^{l-1})$ where $g$ is an activation function such as ReLU ($max(0,\sum_m w_{nm}^l a_m^{l-1}$)). We can rewrite the equation for unit-wise output to represent layer-wise output by vectorizing our components and keeping track of the weights at a given layer in a matrix $W$ which gives $\vartheta^l = g(W^l\vartheta^{l-1})$.} In this way, the resulting estimator $\hat{f}$ outputs results $\hat{f}(x)$ through a process in which successive layers of computational units mathematically transform $\vartheta_{k}$ inputs from proceeding layers such that $\hat{f}(x_i)=\vartheta_{l}\left(\vartheta_{l-1}\left(\ldots \vartheta_{1}\left(x_i ; W_{1}\right) ; W_{2}\right) \ldots ; W_{l}\right)$ where $x_i$ is an input to the network and $W_k$ is a matrix of weights on the connections between layers. 

The network's internal transformations are often referred to as `representations' and the optimization process by which they are arrived at is known as `representation learning' \cite{bengio2013representation}. In this context, `representation' carries a functional meaning. So, to avoid confusion in what follows, I will adopt the term `learned transformations' to refer to a DLM's internal data operations. In \textit{Section 4}, I will consider whether learned transformations can be viewed as representational in the relational sense. The precise transformations at a layer $\vartheta_{k}$ are learned by iteratively updating all weights on all connections, sequentially minimizing a loss function $\mathcal{L}$ through backpropagation of errors \cite{rumelhart1986learning}. The estimator's learned transformations are optimized such that the risk $R$ over the training distribution $P$ is minimized for $R_p(\hat{f}) := \mathbb{E}_{(X,Y)\sim P}[\mathcal{L}(\hat{f}(X),Y)]$.

The mathematical particulars of a deep learning estimator are fully transparent and available to direct inspection \cite{leslie2019understanding}. This fact is sufficient to establish that $\hat{f}$ is representational$_\text{rel}$. To see this, consider that the numerical values of both the input and output layers of the network are taken or interpreted to denote states of a specific target system and are isomorphic to a data-model of those states \cite{suppes1966models,flach2012machine}. Both layers are represented as arrays of values $<a_1, a_2, ... , a_k>$ that correspond to sets of values in their data models $\{x_1, x_2, ... , x_k\}$ which, in turn, correspond with sets of elements in the target system \{‘height in cm’, ‘weight in kg’, ... , ‘points per game\}. From this, it seems unobjectionable that the input and output layers represent targets (and often particular aspects of those targets). So, deep learning estimators are \textit{representational$_\text{rel}$}. Since it has already been established in \textit{Section 2} that deep learning estimators are models, we can now say that they are \textit{representational$_\text{rel}$ models}. 

Having established that deep learning models are representational$_\text{rel}$, we can now turn to the hard question of the degree to which they are semantically decomposable. If DLMs are highly semantically decomposable representations$_\text{rel}$ of their targets, then, in principle, one can assign an interpretation to a large proportion of the units and weights that comprise the network's hidden layers. If, on the other hand, they are holistic models, then, in principle, one cannot assign such interpretations. However, assigning interpretations to a deep learning model's learned transformations is, \textit{in practice}, notoriously frustrated by opacity \cite{lipton2018mythos,raz2022understanding,rudin2019stop,duede2022instruments,duede2022deep}.

DLMs are said to be epistemically opaque \cite{humphreys2004extending}. This means that, following Humphreys, it is impossible for someone to know all of the factors that are epistemically relevant to the model's behavior. The point is perhaps best made by \cite{creel2020transparency}, who identifies three levels of transparency in complex, computational systems: `Algorithmic', `Structural', and `Runtime'.\footnote{See also \cite{zerilli2022explaining} for an account of opacity that turns on notions of intelligibility and fathomability.} DLMs lack algorithmic transparency, as it is impossible to identify the high-level logical rules that govern the learned transformations of inputs to outputs. They also lack structural transparency, rendering it unclear whether or how the distribution of weights and parameterization implements the high-level logical rules. 

As a result, opacity blocks a full assessment of the representational$_\text{rel}$ status of deep learning models for local and global reasons. Locally, because the models are structurally opaque, it is impossible to directly evaluate \textit{how} components (units and weights) of the hidden layers relate to aspects of the target. Globally, because the models are algorithmically opaque, it is impossible to directly evaluate \textit{whether} the components of the hidden layers relate to aspects of the target \cite{sullivan2022understanding}. Of course, this argument does not establish that DLMs are not highly decomposable. Rather, it shows that, due to opacity, one cannot directly establish the representational$_\text{rel}$ status of deep learning models. Instead, one must turn to either theoretical or empirical considerations, which is our task in \textit{Section 4}.

\section{Considerations of Composition and Distribution}

Despite opacity, there may still be empirical, conceptual, and philosophical reasons to believe that deep learning models are more richly representational than minimally semantically decomposable holistic models. The aim of this section, then, is to get clear about what we are and are not justified in believing about whether a deep learning model's learned transformations relate representationally$_\text{rel}$ to aspects of the model's target system. From the preceding arguments, we have seen that 1) the inputs to and outputs from the model can be characterized in terms of the target, and 2) the outputs of the model are a function (a series of learned transformations) of its inputs. Much of the research in explainable AI (XAI) focuses on developing techniques for interpreting learned transformations (representations$_\text{\textbf{func}}$) as a way of making sense of patterns in the model's data operations \cite{raghu2020survey}. 

There are, however, model-directed and world-directed ways of interpreting a model. From a model-directed perspective, the interpretation aims at making sense of how the model arrives at its outputs, as well as how the model would behave under certain perturbations. This is known as mechanistic interpretation. For instance, consider gradient saliency \cite{dumitru2009saliency} applied to an input layer. Here, one calculates a saliency quantity as the gradient of the loss function with respect to changes in the input as a way to evaluate how the model's accuracy is a function of inputs from bounded neighborhoods in feature space. A procedure of this kind allows researchers to make model-directed interpretations of intervals within individual input dimensions that contribute the most to the model's accuracy. Recently, \cite{duede2022deep} has argued that procedures of this kind can, in the context of scientific \textit{pursuit}, give researchers \textit{prima facie} evidence that the corresponding aspects of the target represented by the salient inputs are most relevant for the predicted aspects of the target represented by the outputs.  Crucially, however, such interpretations are local and do not give researchers understanding of how the model reliably relates inputs to outputs globally \cite{raz2022understanding,duede2022instruments}. Indeed, such approaches have come under greater theoretical scrutiny \cite{nie2018theoretical}, are empirically fragile \cite{ghorbani2019interpretation}, and may not be appropriate for tasks requiring a grasp of overall, high-level network logic, such as detecting anomalies, explaining the saliency of the connection between inputs and outputs, or debugging a model \cite{adebayo2018sanity}.

While AI researchers focus on model-directed evaluations of learned transformations, ambiguity in the literature has led some within the AI community and many outside of it to believe that learned transformations are, in fact, representations$_\text{rel}$ or aspects of underlying mechanisms in the \textit{target system} \cite{rudin2019stop,lipton2018mythos,mitchell2019artificial,mitchell2021ai,boge2021two}. In particular, because deep learning models learn their own transformations (representations$_\text{func}$ from data, it is thought that interpretation of these transformations might reveal, as yet undiscovered, mechanisms in the target \cite{samek2019towards,d2019learning}. In this way, the model-directed interpretive aim slips quietly toward a world-directed one.

The remainder of the paper is dedicated to a careful evaluation of some of the key claims in the literature concerning learned transformations in DLMs that give rise to and motivate this slippage. In particular, I will attempt to assess how far these claims can be pressed in the service of supporting the belief that deep learning models are any more semantically decomposable representationally$_\text{rel}$ than, say, the Ising model. While written from the functional perspective, the claims that I will be evaluating are exceedingly suggestive of the view that deep learning models are semantically fine-grained representations of their targets. In particular, I focus on questions of whether the compositional and distributed nature of learned transformations lends support to the idea that DLMs are highly decomposable representations$_\text{rel}$. I show how easy it is for these claims to be heard from the relational perspective but demonstrate how, when carefully examined, they fail to support the belief that deep learning models are any more representational$_\text{rel}$ than has already been established above. While it is surely the case that AI researchers (insiders) are able to sift through the many instances of loose, metaphorical, or otherwise slippery meaning, as I will show, there is no reason to believe that scientists in other disciplines (not to mention philosophers) are always as nimble.

\subsection{Composition}

Consider the following claims. Deep learning models are hierarchical, \textit{composed} of ``multiple levels of representation'' such that information at a given level is transformed ``into a representation at a higher, slightly more abstract level.'' \cite[p.436]{lecun2015deep} And that ``deep architectures can lead to abstract representations because more abstract concepts can often be constructed in terms of less abstract ones.'' \cite[p.1802]{bengio2013representation}. 

If one is not sensitive to the distinction between functional and relational conceptions of representation, it can be tempting to believe that what is being claimed here is that deep learning models encode conceptual primitives at each layer that are then composed into more abstract concepts. If this were true of deep learning models, then, \textit{in principle}, it should be possible to extract from DLMs representations (in the relational sense) of aspects or mechanisms of the target. For example, in \cite{jha2018elemnet} the authors study the performance and learned transformations of a DLM which predicts formation enthalpy of materials. They examine the activations of the network across multiple hidden layers to evaluate how the network is arriving at its predictions. The authors claim to observe that the first layer of the network clearly groups elements based on their group number and period and that later layers represent interactions between elements. Here is the explanation they provide: ``The early layers of the network are learning features based directly on the input values (i.e., presence of certain types of elements)... but it is clear that the later layers are more related to interactions between elements than the presence of single elements. Overall, the activations for both single elements and binary compounds demonstrate the power of deep learning networks to learn essential domain knowledge without specially-designed inputs \cite[p.7]{jha2018elemnet}.'' Citing the quote from \cite{lecun2015deep} above, the authors reason that the network is able to learn domain knowledge in this way because ``each layer of the network is gradually learning more complex features in a way similar to networks built for image classification.'' \cite[p.7]{jha2018elemnet} 

Deep neural networks (particularly ConvNets) certainly pool \textit{information} from one layer into the next. As I demonstrated in \textit{Section 3}, successive layers are mathematical transformations of information in prior layers. Yet, as I argue in this section, the claim that DLMs \textit{compose} salient concepts into something like knowledge of the empirical domain from which the training data are drawn is not justified. The reason for this is that we do not have any good prior reason to believe that the network is, at any level, representing such contents. Rather, that is the assumption we are \textit{starting} with. It is no help to insist that what is composed are `conceptual primitives' or `micro-features' \cite{hinton1984distributed} that are just too complex for humans to intuit, because this claim is in need of as much (if not more) support as the claim that the contents are, in principle, humanly interpretable. After all, if they are not, \textit{even in principle}, humanly interpretable, then what reason would we have for believing that they are there at all? 

Nevertheless, there are at least three distinct approaches to supporting the idea that deep learning models are composed of multiple levels of representation as opposed to mere information transformation. \textit{Approach 1} involves arguing that the deep neural networks provide biologically plausible models of human cognition; \textit{Approach 2} argues that there is compelling empirical evidence that neural networks compose abstract, meaningful concepts from meaningful conceptual primitives; \textit{Approach 3} argues that the nature of composition in DNNs is representation composition and, therefore, representational$_\text{rel}$.

\textit{Approach 1}: One intuition that could motivate the idea that learned transformations are composed into representational$_\text{rel}$ contents comes from our neuroscientific understanding of composition in the human (and primate) visual cortex. Like a deep network, the visual cortex is organized into layers of neurons in which those at one layer `output' their activation magnitudes to neurons at the next layer. Empirical evidence in \textit{neuroscience} supports the conclusion that succeeding layers detect increasingly complex contents \textit{composed} of the elements detected by prior layers in \textit{mammalian brains}. Early layers detect edges or boundaries where there are sharp discontinuities in certain regions. Excitations by edges are passed to higher-layer neurons that are, in turn, excited by stable combinations of edges, and so on until the visual cortex somehow produces in you recognition of an object. Given that DLMs are arranged into `layers' and are feed-forward in much the same way, the thought is that they are likely to behave like the visual cortex. This thinking applies neatly to ConvNets which are also arranged in such that they have `receptive fields' (a region of pixel space values passed to a unit). Indeed, because ConvNets perform so well at vision tasks in comparison to mammalian visual abilities \cite{cadieu2014deep}, it is widely assumed that the best explanation for their accuracy is that they behave rather like the visual cortex. That is, they are arranged and perform in such a way as to be a biologically plausible model of parts of human cognition.

Yet, this would amount to nothing more than question-begging and would be rather like claiming that airplanes can fly because birds also have wings. Philosopher Catherine Stinson has argued persuasively for the implausibility of something like \textit{Approach 1} \cite{stinson2020implausible}. Briefly, there are significant differences between visual processing in the brain and what takes place in ConvNets. In addition to a feed-forward flow of information (which brains and ConvNets `share'), in the visual cortex, there is roughly an order of magnitude \textit{more} feed-backward flow of information. There is no analogous information back-flow in a deep network. Moreover, brains are better thought of as analog devices, not digital, and units in deep nets lack \textit{virtually all} of the dynamics of biological neurons \cite{cichy2019deep,jones2021might}. In \textit{Section 4.2}, I consider the more general posit that, because DNNs are brain\textit{-like}, they must be representational.

\textit{Approach 2}: Salience methods applied to a network's hidden layers can appear to annotate regions of feature space that networks leverage in classification. For example, a correctly classified image of an eagle might generate a saliency map which indicates the salience of pixel values in the areas corresponding to what humans observe to be the eagle's beak, eyes, and the region of high color contrast between its head and torso. The conclusion that researchers might draw from this is that `beak', `eyes', and `white head' are the aspects of the target image that the network learns and relies upon in modeling `eagleness'.

Yet, procedures of this kind are mercilessly underdetermined (a concern shared increasingly within the machine learning community \cite{d2020underspecification,nie2018theoretical,ghorbani2019interpretation,adebayo2018sanity} and the philosophical community \cite{raz2022understanding,raz2022importance,duede2022instruments}). As I argued above, the results of saliency procedures applied to the input layer might conceivably provide insight into how input aspects of the target affect the accuracy of the model on certain tasks. Yet, these local procedures are model-directed ---they tell us something about how regions of input feature space contribute to model accuracy, \textit{not} how aspects of the input target are related to aspects of the output target.

Moreover, the underdetermination of such procedures blocks our ability to empirically evaluate their representational$_\text{rel}$ efficacy. For instance, it could be that the heatmap is rather cool around the eagle but warm around the background indicating that the network has picked up on statistical correlations that have nothing to do with eagles (like blurry backgrounds that tend to accompany wildlife photos \cite{landecker2013interpreting}). This kind of thing, known as shortcut learning, is common, and it is notoriously difficult to determine when it is and is not happening \cite{geirhos2020shortcut,arjovsky2019invariant}. In fact, recent analysis has found that deep ConvNets trained on popular benchmarks overwhelmingly use nonsense, statistical shortcuts that generalize across the data distribution of benchmark images \cite{carter2020overinterpretation}. This `model overinterpretation' is defined to ``occur when a classifier finds strong class-evidence in regions of an image that contain no semantically salient features \cite[pg. 1]{carter2020overinterpretation}'' and that this `pathology' applies equally to adversarially robust models.

Even if we make a strong assumption that a model is not keying on statistical shortcuts, visual inspection of saliency maps is still highly underdetermined. For instance, it could be that \textit{inactivity} of certain units is as, if not more, important to classification. After all, that $p$ is \textit{not} $q$ is informative. In fact, \textit{Approach 2} and \textit{Approach 1} both suffer from empirical counter examples drawn from evidence that inactive or low-active neurons in the mammalian brain are informationally salient \cite{jones2021might}.\footnote{Adversarial examples in image classification also serve as strong counter examples against \textit{Approach 2}. Here, humanly imperceptible perturbations made to the pixel values of otherwise correctly machine-classified images induce astonishingly inaccurate classifications for which the network is, nevertheless, extremely confident \cite{nguyen2015deep}.}

\textit{Approach 3}: As we saw in \textit{Section 3}, deep learning models are transitive, compositional functions. One might be tempted to think that the formal properties of neural network function composition give us a reason to think that they encode representations$_\text{rel}$ at each layer. The thinking is something like ``deep learning really is a \textit{representation$_\text{rel}$}-learning method in which concepts go in, are represented as functions which are further composed, recombined and assembled, and higher-level, abstract concepts come out.'' As Buckner puts it, the abstract concepts that come out are ``hierarchically composed all the way down, from pixels and color channels to lines, angles, and shape'' and that formal aspects of deep networks serve to ``detect and compose each layer of features into those at the next layer of abstraction.'' \cite[pg. 5347]{buckner2018empiricism}

The AI literature tends to use the terms `layer' and `level' interchangeably which facilitates an easy association (and \textit{conflation}) with notions of levels of organization. This, in turn, has both metaphysical and ontological consequences for how we interpret what is happening \textit{layer-wise} within a neural network. The compositional thesis leverages the metaphor that layers are \textit{levels} of organization and, moreover, since there is a lot of function composition going on in a DLM, the connection to abstract concept composition is ready-made \cite{fodor1988connectionism}. Thinking of this kind can be seen playing out in the materials science paper cited above.

To say that a representation at a given level is a ``transformed representation'' of the previous level is, presumably, to say something like layer $l_n$ is a representation of $l_{n-1}$. In the same way, $l_1$ would be a representation of $l_0$. This means that each layer is, itself, a representation, and, moreover, it is a representation of a representation. If successive layers are representations of prior layers, we must conclude that $l_n$ (the final layer) is a representation of $l_0$ (the initial layer). We already saw that the output of a deep network is a highly nonlinear transformation of the input. However, if we accept this kind of approach, it still does not establish that individual layers represent$_\text{rel}$ aspects of the target. Rather, it establishes that succeeding layers represent$_\text{func}$ preceding layers.

There is, however, the possibility of \textit{informational transitivity} such that, if $B$ represents $A$, and $C$ represents $B$, then information about $A$ will be present in the representation $C$. If we consider a photograph of a painting of an orchard that contains the whole painting, then the visual information encoded in the painting is, for the most part, preserved. Nevertheless, the photograph is a representation$_\text{rel}$ of \textit{the painting}, not the orchard. This process can be repeated by, say, painting a picture of the photograph and then photographing the new painting ---on and on. Presumably, specific informational signals such as locality will be preserved such that, if a pixel in the second photograph is red, then an adjacent, contiguous pixel will be red with high probability. If that pixel is not red, then this region in pixel space likely corresponds to a boundary between one object and another. This provides some \textit{prima facie} reason to believe that the contents of the original painting are informationally present in the final photograph. Extending this line of reasoning to the flow of information through a deep neural network \textit{seems} reasonable, particularly considering that the output layer is reached by recursively transforming recursively transformed vectors. However, adversarial examples provide good reason to doubt that this so-called `smoothness prior', or locality generalization, is guaranteed. In fact, informational signals are not reliably stable throughout DLMs \cite{szegedy2013intriguing}. In this sense, not only is function composition distinct from semantic composition, but, function composition can lead to \textit{semantic loss}. Strangely, non-robust models have proven to be very effective for prediction, raising the possibility that semantic coherence and predictive efficacy do not go hand in hand \cite{tsipras2018robustness,ilyas2019adversarial,zhou2019humans}. Besides, even if the smoothness prior \textit{did} hold, this would be consistent with a functional conception of representation where an extension of the concept to a relational conception would still stand in need of justification. 

From what we have seen, considerations of composition fail to provide us with a good reason to believe that any given DLM encodes a highly, semantically decomposable representational$_\text{rel}$ model of its target. I turn now to considerations of distribution.

\subsection{Distribution}

That deep learning models are `distributed' is often leveraged to explain why semantic interpretations of the model are difficult to divine (though it just as easily could account for their holistic nature). Their distributed nature provides \textit{prima facie} reason to believe that \textit{if} DLMs encode semantically decomposable representations$_\text{rel}$ of their targets, \textit{then} these representations$_\text{rel}$ will be difficult to suss out since they are distributed throughout an often astonishingly extensive network. This is an example of what Creel refers to as structural opacity. Here, the high-level logic that governs the transformation of input to output, describable in terms of the target system, is inaccessible due to how that logic is realized in the network \cite{creel2020transparency}. Of course, given the preceding discussion, we have encountered no good reason to believe the antecedent of this conditional.

Nevertheless, it is worth considering whether distribution yields good reasons to believe that DLMs encode highly decomposable semantic representations$_\text{rel}$ of their targets \textit{independent} of considerations of composition. The concept of distribution allows for two distinct \textit{Tacks} to accomplish this, each based on a slightly different way of understanding the distribution thesis.

\textit{Tack 1}: This maneuver argues from a particular interpretation of the idea of distributed transformations expressed in slightly different terms by Hinton and Bengio. Hinton's version: ``A distributed representation uses a unit for a set of items, and it implicitly encodes a particular item as the intersection of the sets that correspond to the active units.'' \cite[p.94]{hinton1984distributed} Bengio's version: ``\dots each concept is represented by having $k$ features  being  turned  on  or  active,  while  each  feature  is  involved  in  representing many concepts.''\cite[p.1801]{bengio2013representation} To see how \textit{Tack 1} interprets these claims, it is helpful to consider an example of a distributed representation that fits the interpretation. Biological taxonomic trees are distributed representations because an organism's \textit{Family} can be represented as the \textit{intersection} of more general or abstract taxonomic ranks, each of which participates in representing many families. Since a highly semantically decomposable representational$_\text{rel}$ model is such that one can specify in a regular and systematic manner what the constituents of the model mean, interpreting the distribution thesis along the lines of \textit{Tack 1} gives us a clear path to shore. Here, the constituents of the distributed representation are the individual units, and, in principle, the interpretation of the model can be carried out in the same manner as that of the taxonomic tree. The difficulty, when it comes to interpreting DLMs, according to \textit{Tack 1}, is that, with many units, determining the sets of contents that they represent$_\text{rel}$ is daunting. In this way, perhaps distribution gives us both a reason to believe that DLMs are semantically decomposable representations$_\text{rel}$ of their targets \textit{and} explains why they are structurally opaque. 

Yet, \textit{Tack 1} is a nonstarter because it is simply a restatement of the argument from composition, yet, this time, in terms of distribution. As we saw when considering composition (\textit{Section 4.1}), to proceed under the assumption that individual units represent$_\text{rel}$ in the way proposed by \textit{Tack 1} is just question-begging. However, one might argue that \textit{Tack 1} does not require a compositional thesis because the individual \textit{layers} of the network are distributed representations$_\text{rel}$. Here, each layer is a vector of units, and the orthogonality of individual units is analogous to the parameters in a linear model such that the weights on a given unit encode the relative contribution of the content \textit{represented$_\text{rel}$ by that unit} to the output \cite{girshick2014rich} which is the entire next layer. However, this claim simply pushes the compositional thesis down to the intra-layer level and does nothing to justify the claim that individual \textit{units} have specifiable contents in terms of the system ---it assumes it. Moreover, on this interpretation, DLMs appear to be \textit{local} rather than distributed schemes.

\textit{Tack 2}: This tack argues for a much stronger interpretation of distribution than \textit{Tack 1}. Here, distribution is interpreted the as implying that representational$_\text{rel}$ contents are \textit{fully distributed} throughout the network such that ``each [unit] is involved in representing many different [contents]'' means \textit{every} weight on \textit{every} unit is involved in computing the representation$_\text{rel}$ of \textit{every} content. Given that every unit is ``active'' in representing$_\text{rel}$ every entity (regardless of whether its output is zero), only the entire function can be said to represent$_\text{rel}$ any \textit{particular} content. Varying a weight \textit{anywhere}, distributes changes to the representation$_\text{rel}$ \textit{everywhere}. A familiar and perhaps more intuitive example of such distributional effects comes from the mathematics of Fourier transforms, whereby a modification of $f$ anywhere distributes changes to $\hat{f}$ everywhere.

It is worth pointing out that a view like the one motivated by \textit{Tack 2} is certainly consistent and compatible with a \textit{functional} understanding of representation. That is, there is no reason to think that a model's learned transformations cannot be fully distributed in this way (which might be one reason why widespread transfer learning has proved so difficult). However, if we have a distributed \textit{(relational)} representation of some aspects of the model's target, where might we look for the assignable meaning? For any given content, the content of that representation$_\text{rel}$ would be distributed throughout all units and innumerable connections. So, it isn't any wonder that digging out fine-grained contents from DLMs has proved impractical. In this way, \textit{Tack 2} explains why we struggle to recover mechanisms, knowledge, and other semantically decomposable contents from DLMs.

But, how does this interpretation support the antecedent claim that the network encodes semantically representational$_\text{rel}$ contents of its target? It runs into immediate conceptual difficulty. Far from giving us good reason to believe that the DLMs are highly decomposable representational$_\text{rel}$ models, \textit{Tack 2} gives us good reason to believe that the model's contents must be either highly contingent or completely arbitrary. Given that, in a distributed representation, each unit is involved in representing \textit{every} content, there is certainly no \textit{a priori} reason to suspect that individual units would encode \textit{discrete}, semantically meaningful contents. After all, every unit is sensitive to every parameter in the input space and, unlike local representational schemes, in distributed representations, a change in the value of a single weight (which would be affected by either adding or subtracting a single sample to the training data, changing the number of training epochs, adjusting the learning rate, adding a layer, or not changing anything and simply doing the whole training again), causes a change in the sensitivity and learned transformations everywhere such that the contribution made by each of the individual units to relating to every content in the target system is modified. Given that unit-wise contribution is globally modifiable by means of distinct modifications to individual weights, there can be no good conceptual reason to believe that individual units encode specifiable semantic content because that content is far from invariant and highly contingent. If anything, this is an argument in favor of holism about decomposability.

To resolve this difficulty, \textit{Tack 2} must appeal to the idea that fully distributed representation$_\text{rel}$ allows for a plausible analogy between learned transformations in DLMs and functional representations in brains. After all, we have reason to believe that a brain encodes models of its environments, these models are distributed in the brain's neural architecture, and we have a tough time finding them in neural correlates \cite{shagrir2018brain}. Since we see this kind of phenomenon in highly distributed machines like brains, we should expect them in similar machines. That is, one might insist, ``this is how brains work. Surely, you do not want to deny that brains encode distributed representations!''. 

However, there are several problems with this approach (some of which we have encountered already). First, the way highly semantically decomposable models represent$_\text{rel}$ the world and how the world is represented in the brain are very different. While it is true that brains are distributed systems in which specific neurons have been observed to be sensitive to particular contents in the world, that is a) consistent with a \textit{functional} conception of representation, and b) a far cry from believing that a representational$_\text{rel}$ model of the world could be \textit{read off} of the neural architecture of a human brain. The neuroscientific community abandoned such thinking in the early 2000s. While we have good reason to believe that the brain encodes highly representational models, this is not because we have observed them in brains. Studies of neural correlates are notoriously under-determined in much the same way as the salience methods discussed above. Instead, brains routinely make their representations explicit to consciousness through various cognitive and communicative acts. In fact, it has recently been argued that representational$_\text{rel}$ encodings of the sort that one might seek to locate in DLMs are both logically and empirically in conflict with the causal structure and activity of the brain \cite{brette2019coding}. As Rumelhart, the famous cognitive scientist and AI pioneer, once said, ``information is not stored anywhere in particular. Rather it is stored everywhere. Information is better thought of as `evoked' than `found'.'' \cite{rumelhart1986general}. Even if certain units appear to \textit{correlate} with certain determinate features of reality, the semantic content of that unit remains highly under-determined. That the brain has evolved a means to transcribe \textit{its} models is certainly not a good reason to believe that DLMs encode such models. Finally, Hinton himself admits that given the nature of distributed representation, units could ``correspond to conceptual primitives, or they may have no particular meaning as individuals''\cite[p. 33]{rumelhart1986general}. In fact, as I have shown, there is no good reason why they should.

Writing more than three decades before Rumelhart, Wittgenstein quipped that ``nothing seems more possible to me than that people some day will come to the definite opinion that there is no copy in the \dots nervous system which corresponds to a \textit{particular} thought, or a \textit{particular} idea, or memory.'' \cite{wittgenstein1996last} And, we have no good reason to think that the same must not be said of deep learning models. \textit{Tacks 1} and \textit{2}, then, are fruitlessly beating to windward.

Throughout \textit{Sections 3} and \textit{4}, I have argued that the nature of deep learning models provides us with no good reason to believe that they encode highly decomposable representations$_\text{rel}$ models of their targets. If, in the course of a scientific investigation, a DLM is examined in such a way that a scientist infers an explanation for its predictions in terms of relations that hold between features of the target system, then we need good reasons (justification) for believing that what is inferred is, in fact, encoded in the network, and these reasons must be independent of the belief that DLMs are highly decomposable representational$_\text{rel}$ models. But, no such justification is forthcoming. This has been my claim throughout. As a result, I claim that deep learning models are best conceived as type of holistic models, though situating deep learning in that wider class is a topic for future analysis.

As to their degree of idealization, this cannot be evaluated globally, as each trained model is unique in both its target and functional representation. However, it does seem plausible that any given model might fall anywhere along the idealization dimension, with some models representing targets in a highly idealized manner while others doing so in a more minimally idealized way. Determining the degree of idealization of any given model is, however, blocked by opacity.

\section{Conclusion}
Deep learning has become increasingly central to science, in large part due to its capacity to quickly, efficiently, and accurately predict and classify phenomena of scientific interest. It is common for both scientists and deep learning researchers to refer to DLMs as having encoded representational models of their targets. Research into understanding and interpreting these models proceeds under a pair of assumptions. The first is that DLMs functionally represent data in a way that is conducive to accurate prediction and classification. These learned transformations of data can be interrogated in manner that allows for some local understanding of how aspects of the input data contribute to the model's overall accuracy. This model-directed assumption, though perhaps empirically dubious, is justified by a conception of representation that is mechanistic and functional in nature \cite{cao2022putting}. The second assumption, however, takes learned transformations to represent aspects of the target system in a semantically decomposable and relational sense. On \textit{this} assumption, it is possible to, in principle, interpret the model's learned transformations in a world-directed manner to learn about aspects, relations, mechanisms, or other features of the target system. If this were the case, then successful interpretation could furnish scientists with higher and more sophisticated levels of scientific understanding about the world. Yet, the arguments of this paper have sought to demonstrate that, in general, we have no good reason to believe that DLMs are locally semantically decomposable representations$_\text{rel}$ of their targets. At most, we are justified in classifying them as holistic representational$_\text{rel}$ models. 

This result exposes a potential epistemic risk associated with attempts to semantically interpret DLMs as representational$_\text{rel}$ models in a world-directed manner, as such attempts suffer from being either highly under-determined and unfalsifiable. This means that, as of now, the direct interpretation of deep learning model parameters for scientific as explanations is not epistemically justified. However, and importantly, this does not mean that deep learning cannot aid in furthering scientific discovery \cite{duede2022deep}. For instance, predictions made on the basis of DLMs can serve as data used for the evaluation, extension, or even knowledge of certain theories \cite{devries2018deep,duede2024apriori}, supply researchers with novel intuitions concerning the behavior of specific systems \cite{wang2019massive}, or simply bypass computational roadblocks on the way to discovery \cite{senior2020improved}. In short, while we have no good reason to believe that deep learning models stand in highly decomposable representational$_\text{rel}$ relations to the world, this does not mean that their behavior \cite{rahwan2022machine} and astonishing predictive capacities cannot be leveraged as a means to deeper, more fundamental scientific understanding \cite{wimsatt1987false}.

\bibliographystyle{alpha}
\bibliography{bibliography}

\end{document}